\documentclass{article}

\usepackage[nonatbib,preprint]{neurips_2021}

\usepackage[utf8]{inputenc} 
\usepackage[T1]{fontenc}    
\usepackage{hyperref}       
\usepackage{url}            
\usepackage{booktabs}       
\usepackage{amsfonts}       
\usepackage{nicefrac}       
\usepackage{microtype}      
\usepackage{xcolor}         

\usepackage{xparse}
\usepackage{ifthen}

\usepackage{amsmath,amssymb,amsfonts}
\usepackage{mathtools}
\usepackage{dsfont}
\usepackage{amsthm}
\usepackage{thmtools, thm-restate}

\usepackage{pgfplots}
\usepackage{pgfplotstable}
\usetikzlibrary{spy}
\usepgfplotslibrary{groupplots}
\pgfplotsset{compat=1.16}

\usepackage{booktabs}
\usepackage[section]{placeins}
\usepackage{wrapfig}

\usepackage{hyperref}
\usepackage{siunitx}
\usepackage{todonotes}
\usepackage{biblatex}
\usepackage{enumitem}
\usepackage{xspace}
\usepackage{csquotes}
\usepackage{multicol}

\usepackage{siunitx}
\usepackage{float}

\DeclareMathOperator{\expectsymb}{\mathds{E}}
\DeclareMathOperator*{\argmin}{\mathrm{argmin}}

\newcommand{\indsymb}{\mathds{1}}
\newcommand{\probsymb}{\mathds{P}}
\newcommand{\reals}{\mathds{R}}

\newcommand{\transposed}{^\mathsf{T}}

\let\set\undefined
\NewDocumentCommand{\set}{m}{\left\{#1\right\}}
\NewDocumentCommand{\abs}{m}{\left|#1\right|}

\NewDocumentCommand{\wrapbrackets}{m}{\left[ #1 \right]}

\NewDocumentCommand{\expect}{e{_}o}{
    \def\decorated{\expectsymb\IfNoValueTF{#1}{}{_{#1}}}
    \IfNoValueTF{#2}{\decorated}{\decorated\!\wrapbrackets{#2}}
}
\NewDocumentCommand{\trisk}{e{_}o}{
    \def\decorated{\operatorname{R}\!{}^*\IfNoValueTF{#1}{}{_{#1}}}
    \IfNoValueTF{#2}{\decorated}{\decorated\!\wrapbrackets{#2}}
}

\NewDocumentCommand{\terisk}{e{_}o}{
    \def\decorated{\operatorname{\hat{R}}\!{}^*\IfNoValueTF{#1}{}{_{#1}}}
    \IfNoValueTF{#2}{\decorated}{\decorated\!\wrapbrackets{#2}}
}

\NewDocumentCommand{\orisk}{e{_^}o}{
    \def\decorated{\operatorname{R}\IfNoValueTF{#1}{}{_{#1}}\IfNoValueTF{#2}{}{^{#2}}}
    \IfNoValueTF{#3}{\decorated}{\decorated\!\wrapbrackets{#3}}
}

\NewDocumentCommand{\oerisk}{e{_^}o}{
    \def\decorated{\operatorname{\hat{R}}\IfNoValueTF{#1}{}{_{#1}}\IfNoValueTF{#2}{}{^{#2}}}
    \IfNoValueTF{#3}{\decorated}{\decorated\!\wrapbrackets{#3}}
}

\NewDocumentCommand{\ind}{o}{
    \IfNoValueTF{#1}{\indsymb}{\indsymb\!\set{#1}}
}
\NewDocumentCommand{\prob}{o}{
    \IfNoValueTF{#1}{\probsymb}{\probsymb\!\set{#1}}
}

\NewDocumentCommand{\vect}{m}{\mathbf{#1}}

\NewDocumentCommand{\gradient}{o}{\IfNoValueTF{#1}{\nabla}{\nabla_{\!#1}}}
\NewDocumentCommand{\hessian}{o}{\IfNoValueTF{#1}{\nabla^2}{\nabla^2_{\!#1}}}

\pgfplotstableset{
BasicTableStyle/.style={
    every head row/.style={
        before row={\toprule},
        after row={\midrule}
    },
    every last row/.style={after row=\bottomrule}
}}

\makeatletter
\DeclareRobustCommand\onedot{\futurelet\@let@token\@onedot}
\def\@onedot{\ifx\@let@token.\else.\null\fi\xspace}
\newcommand{\ie}{i.e\onedot}
\newcommand{\eg}{e.g\onedot}

\newcommand{\cf}{cf\onedot}

\makeatother

\theoremstyle{definition}

\theoremstyle{remark}

\pgfplotstableset{
  TableWithRules/.style={
    every head row/.style={
        before row={\toprule},
        after row={\midrule}
    },
    every last row/.style={after row=\bottomrule},
  }
}

\newcommand{\dataset}{\mathsf{X}}
\newcommand{\positives}{\mathsf{P}}
\newcommand{\negatives}{\mathsf{N}}
\newcommand{\posmean}{\bar{\vect{p}}}
\newcommand{\negmean}{\bar{\vect{n}}}
\newcommand{\ftrmean}{\bar{\vect{x}}}

\newcommand{\instance}{\vect{x}}
\newcommand{\labels}{\vect{y}}
\newcommand{\labelcmp}{y}

\newcommand{\numinstances}{n}
\newcommand{\numlabels}{l}
\newcommand{\numfeatures}{d}

\newcommand{\instancespace}{\mathcal{X}}
\newcommand{\labelspace}{\mathcal{Y}}

\newcommand{\weightvec}{\vect{w}}
\newcommand{\init}{\weightvec_0}

\newcommand{\lham}{l_{\text{H}}}
\usepackage{todonotes}

\title{Speeding-up One-vs-All Training for Extreme Classification via Smart Initialization}

\author{%
  Erik Schultheis \\
  Department of Computer Science\\
  Aalto University\\
  Helsinki, Finland \\
  \texttt{first.last@aalto.fi} \\
  \And
  Rohit Babbar \\
  Department of Computer Science\\
  Aalto University\\
  Helsinki, Finland \\
  \texttt{first.last@aalto.fi}
}

\begin{document}

\maketitle

\begin{abstract}
  In this paper we show that a simple, data dependent way of setting the
  initial vector can be used to substantially speed up the training of linear
  one-versus-all (OVA) classifiers in extreme multi-label classification
  (XMC). We discuss the problem of choosing the initial weights from the
  perspective of three goals. We want to start in a region of weight space
  a) with low loss value, b) that is favourable for second-order optimization,
  and c) where the conjugate-gradient (CG) calculations can be performed
  quickly. For margin losses, such an initialization is achieved by
  selecting the initial vector such that it separates the mean of all
  positive (relevant for a label) instances from the mean of all negatives
  -- two quantities that can be calculated quickly for the highly imbalanced
  binary problems occurring in XMC. We demonstrate a speedup of $\approx
  3\times$ for training with squared hinge loss on a variety of XMC
  datasets. This comes in part from the reduced number of
  iterations that need to be performed due to starting closer to the
  solution, and in part from an implicit negative mining effect that
  allows to ignore easy negatives in the CG step. Because of the convex nature
  of the optimization problem, the speedup is achieved without any
  degradation in classification accuracy.
\end{abstract}

\section{Introduction}

\paragraph{Extreme Classification}
In this work we consider \emph{extreme multi-label classification} (XMC)
problems. Here, the number of labels $\numlabels$ is very large (possibly in
the millions). Such problems arise in various domains such as annotating large
encyclopedia \cite{partalas2015lshtc}, image-classification
\cite{deng2010does}, next word prediction
\cite{mikolov2013efficient,mnih2008scalable}, as well as recommendation
systems, web-advertising and prediction of related searches \cite{Agrawal13,
prabhu2014fastxml, jain2019slice}.

 We assume that the fraction of positive training instances for most labels is
very low -- the label frequency has a long-tailed distribution
\cite{pmlr-v9-dekel10a}. This is a good approximation \eg for the
large dataset from the extreme classification repository \cite{Bhatia16} (\cf Figures 1 in
\cite{Qaraei_et_al_2021,dismec,partalas2015lshtc} for some examples),
and for many types of
data that is gathered at internet-scale \cite{adamic2002zipf}.

\paragraph{One-vs-All Classifiers}
Many XMC methods employ some form of a \emph{one-vs-all} (OvA) classifier as
the last stage of the classification procedure, in the sense that for a data
point $\instance$ with corresponding features $f(\instance)$ and label $y \in
\set{-1, 1}$, the predictions are calculated as
\begin{equation}
     \operatorname{top}_k \{ \vect{w}_i\transposed f(\instance) : i \in [\numlabels] \}
 \end{equation}
and the training objective is to minimize (with some additional regularization
terms)
\begin{equation}
   \phi( y_i \vect{w}\transposed f(\instance_i)),  
\end{equation}
where $\phi$ denotes some margin loss and $y_i$ is the corresponding label.
Algorithms that follow this general structure are DiSMEC \cite{dismec} and
ProXML \cite{proxml} for sparse representations $f$, and Slice
\cite{jain2019slice} where $f$ is some pre-trained mapping of instances to
dense vectors. Other examples are Parabel \cite{prabhu2018parabel} and
PPDSparse \cite{yen2017ppdsparse}. Embedding-based approaches where the
mapping $f$ itself is learnt, such as XML-CNN
\cite{Liu2017DeepLF,NEURIPS2019_9e6a921f} are incompatible with the method
presented in this paper. However, in some cases, even though $f$ is not
parameter-free, it is only fine-tuned during the OvA training, as is the case
in Astec \cite{dahiya2021deepxml}.

\paragraph{Fast Training}
Minimizing this objective using a gradient-based iterative algorithm is very
computation-intensive. To reduce compute requirements -- and thus energy consumption -- we can make use of three generic principles
\begin{enumerate}
    \item Try to achieve more progress per step, e.g. by using a second-order method \cite{dismec,keerthi2005modified,galli2021study} or 
    an adaptive step size. \cite{JMLR:v12:duchi11a,ruder2017overview} 
    \item Make the computations of each step faster, e.g. by approximating the true loss
    with negative mining \cite{jain2019slice,dahiya2021deepxml,prabhu2018parabel,yen2017ppdsparse,reddi2019stochastic}.
    \item Reducing the number of necessary steps by taking a good initial guess for the weight vector. \cite{fang2019fast,keerthi2005modified} 
\end{enumerate}
In this work, we focus on the second-order optimization approach using a
truncated conjugate-gradient (CG) Newton optimization, as it is used in recent
versions of Liblinear \cite{galli2021study} (older versions
\cite{fan2008liblinear} were using a trust-region Newton method).

\paragraph{Implicit Negative-Mining with Hinge-Like Losses}
To a certain degree, the second point is achieved implicitly when using a loss
function that becomes zero once the point is classified correctly with
sufficient margin. In that case, such a training point makes no contribution
to the gradient or Hessian, and thus can be skipped in these computations.
This can be seen as an implicit negative mining, where the list of negatives
is updated each iteration (similar to \textcite{yen2017ppdsparse}) based on
the values of $y_i \vect{w}\transposed f(\instance_i)$. Unlike explicit
negative mining, this is not an approximation, but it also does not enforce
sparsity but only exploits it when it exists. Fortunately, close to
convergence when most instances are correctly classified, this leads to very
sparse computations for the CG procedure. Thus one of our goals is to get into
this fast regime as quickly as possible.

\paragraph{Choice of Initial Vector}
The third idea is used in Liblinear to speed up hyperparameter sweeps. It is
assumed that the solutions for two similar hyperparameter values are closely
related, and as such the final weight vector of one training run can be used
to warm start the next. The feasibility of this approach in (cold-start) XMC
has been shown in \textcite{fang2019fast}. The basic motivation is given by
their Lemma 3.1, which states that if two label vectors are similar (measured
in Hamming distance), then the optimal weight vectors with respect to a
Lipschitz loss are also similar.

The strategy is thus to train the optimal weight vector for a virtual label
that has only negatives. Such a label vector has small Hamming distance to all
the label vectors of tail labels, and thus provides a good candidate for a
starting vector. They call this strategy \emph{OvA-Primal} (OvAP).

They extend this idea by building a \emph{minimum spanning tree} (MST)
over the labels (plus the virtual all-negative label at the root), and then
training in such an order that during each training process the trained
weights of a label that is very similar to the target label can be reused.
This is called OvAP++.

\paragraph{Shortcomings of the Existing Work}
OvAP is very simple and easy to implement, but it completely ignores
the positive labels for each instance. We will show that there is a
computationally cheap method that allows taking them into account.

In contrast, OvAP++ makes use of the positives, but requires much more care to
implement, as it breaks the embarrassingly parallel nature of the original OvA
approach. Now the individual binary problems need to be solved in a certain
(partial) order. This also precludes the application of this strategy to
situations where all labels have to be trained simultaneously, \eg when
fine-tuning the feature representations $f$.

The evaluation of that strategy has not been done on datasets which are very
large and show extreme tail-label behaviour (Wiki31K is the most "extreme"
dataset they have used). In such a situation, the speed-up due to the tree
procedure might not be very large, because the tail-labels will not have
overlapping positives. This is supported by \cite[Lemma 3.2]{fang2019fast},
which implies that two labels with non-overlapping positives need not be
considered for the MST. Since the root node is fixed by construction to be the
all-zero label, this is likely to lead to a graph for which most of the nodes
(corresponding to the tail labels) are direct children of the root, and are
trained exactly as they would without the MST.

Important properties of the loss function (such as local smoothness /
Lipschitz constant) sometimes are more favourable after some training epochs
than right at initialization time, as evidenced \eg by the success of cyclical
learning rates \cite{smith2017cyclical} and warmup phases
\cite{goyal2018accurate}. However, the existing analysis focuses only on
finding an initial weight vector that has loss as low as possible.

\paragraph{Contributions}
In this paper, we discuss why considering the loss value at the intitial
weights alone is not a sufficient criterion for selecting initialization
strategies, and discuss which other properties are necessary
(\autoref{sec:analysis}). Based thereon, we present a novel initialization
strategy that keeps the simplicity of OvAP, but incorporates information about
the positive instances for each label to find even better initial weights
(\autoref{sec:aop-init}). We extend the empirical evaluation of
\textcite{fang2019fast} to cover larger datasets with up to 3 million labels,
and demonstrate that the proposed method works for both sparse and dense
feature spaces (\autoref{sec:experiments}).

We show that, when employed in conjunction with the squared-hinge loss
function, the benefit of this initialization is not limited to just starting
closer to the final weight vector, but also leads to much faster computations
in each training step because of the implicit negative mining. In the
converse, we present an example where a certain choice of initial vector
provides a decrease of initial loss by several orders of magnitude, but still
results in no significant improvement in training time.

\section{Conjugate-Gradient Newton Optimization}
\paragraph{Setup}
We are given a dataset $\mathcal{D} = \set{(\instance_i,
\labels_i): i \in [\numinstances]}$ with $\numinstances \coloneqq |\dataset|$
training instance $\instance \in \instancespace = \reals^\numfeatures$ and their
corresponding labels $\labels \in \labelspace = \set{-1,
1}^\numlabels$. For a label $j \in [\numlabels]$ we denote with $\positives(j)
\coloneqq \{\instance_i:
\labelcmp_{ij} = 1\}$ the instances for which the label is relevant, and analogously 
$\negatives(j) = \dataset \setminus \positives(j)$, where $\dataset \coloneqq
\set{\instance_i: i \in [\numinstances]}$ is the set of all training
instances. This corresponds to a weight vector $\labels^{(j)}$ which has
entries $1$ for each positive instance, $y^{(j)}_i = 1 \Leftrightarrow i
\in \positives(j)$.

We assume that the number of labels $\numlabels$ is large enough that a
preprocessing step of $O(\numinstances)$ does not disrupt the computational
budget, and that the labels are sparse, in the sense that for most $j \in
[\numlabels]$ we have $|\positives(j)| \ll |\negatives(j)|$. These assumptions
are part of what characterizes an XMC problem.

The goal is to find a weight vector $\weightvec_j^*$ for each label $j \in
[\numlabels]$ that minimizes the risk of the corresponding binary problem with
some convex margin loss $\phi$, combined with a regularizer $\mathcal{R}$:
\begin{align}
    \mathcal{L}[\weightvec, \labels^{(j)}] &\coloneqq \sum_{i=1}^{\numinstances} \phi(y^{(j)}_{i}\weightvec\transposed \instance_i) \\
    \weightvec^*_j \coloneqq \weightvec^*(\dataset, \labels^{(j)}) &\coloneqq \argmin_{\weightvec} \mathcal{L}[\weightvec, \labels^{(j)}] + \mathcal{R}[\weightvec].
\end{align}
Throughout this paper, we have used the $L_2$ norm as the regularizer. From
hereon out we will drop the superscript $\labels^{(j)}$ and write $\labels$
for the vector of labels values for a given label.

\paragraph{Optimization Step}
The minimization is carried out using a CG Newton procedure inspired by
Liblinear \cite{fan2008liblinear,galli2021study}. There, a descent
direction $\vect{p}$ is determined by minimizing a locally-quadratic approximation 
$\hat{\mathcal{L}}$ to the loss function
\begin{equation}
    \hat{\mathcal{L}}(\weightvec + \delta \weightvec) = \mathcal{L}(\weightvec) + \nabla \mathcal{L} \cdot \delta \weightvec + \frac{1}{2} \delta \weightvec\transposed \mathsf{H}_{\mathcal{L}} \delta \weightvec,
\end{equation}
where $\mathsf{H}$ denotes the Hessian. Then a line-search using a
backtracking approach is carried out over $\weightvec + \lambda \vect{p}$,
$\lambda \in (0, 1]$. The search direction $\vect{p}$ can be determined
efficiently using CG procedure, as that does not need the full Hessian, but
only Hessian-vector products which can be calculated
as\cite{keerthi2005modified}
\begin{equation}
    \mathsf{H} \vect{d} = \sum_{i=1}^n \phi^{\prime\prime} (\instance_i^\mathsf{T} \vect{w} \cdot y_i) y_i \langle \instance_i, \vect{d} \rangle \cdot \instance_i.
\end{equation}

\paragraph{Squared Hinge Loss}
As the margin loss $\phi$ we use the squared hinge loss (following \eg
\cite{jain2019slice,dismec,proxml})
\begin{equation}
    \phi(m) = \max(0, 1 - m)^2.
\end{equation}
Even though this function's second derivative does not exist at $m=1$, it can
be used for second-order optimization\cite{galli2021study}. Its hinge
structure means that for $m > 1$ we have $\phi(m) = \phi^\prime(m) =
\phi^{\prime\prime}(m) = 0$, which implies that the sum over all data points
can be replaced with a sum over implicitly-mined hard instances $\mathcal{A} =
\set{i: \instance_i^\mathsf{T} \vect{w} \cdot y_i < 1}$\cite{keerthi2005modified},
\begin{equation}
    \mathsf{H} \vect{d} = \sum_{i \in \mathcal{A}} \phi^{\prime\prime} (\instance_i^\mathsf{T} \vect{w} \cdot y_i) y_i \langle \instance_i, \vect{d} \rangle \cdot \instance_i,
\end{equation}
which can be computationally much more efficient than the full sum. The active
set $\mathcal{A}$ has to be re-calculated after each weight update, but each
weight update requires multiple $\mathsf{H} \vect{d}$ calculations for the
conjugate-gradient procedure.

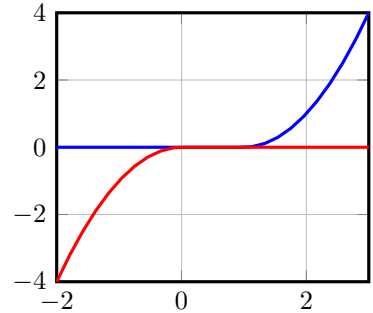
\begin{wrapfigure}[18]{r}{0.33\linewidth}
\pgfplotsset{
    /pgf/declare function={
        phi(\x) = (\x < 1) * (1-\x)^2;
    },
}
\begin{tikzpicture}
\begin{axis}[very thick, width=0.9\linewidth, enlargelimits=false, no markers, scale only axis, grid]
    \addplot+[domain=-2:3] {(1-x)^2 - phi(x)};
    \addplot+[domain=-2:3] {-phi(x + 1)};
\end{axis}
\end{tikzpicture}
\vspace{-1.5em}
\caption{Approximation errors $e(\delta; m_0)$
for $m_0 = 0$ (blue) and $m_0=1$ (red) over $\delta$.}
\label{fig:approx-error}
\end{wrapfigure}

In order for the CG Newton optimizer to make good progress, the loss
$\mathcal{L}$ should be well approximated by its second-order Taylor
expansion. As $\phi$ is piecewise-quadratic, this is fulfilled when $m$ is far
from the boundary at $m=1$. Let the quadratic approximation $\hat{\phi}_{m_0}$
be defined as
\begin{equation}
    \hat{\phi}_{m_0}(\delta) = \phi(m_0) + \delta \left. \phi^{\prime} \right|_{m_0}  + 0.5 \delta^2 \left. \phi^{\prime\prime} \right|_{m_0},
\end{equation}
then we can calculate the approximation error based on the step size as
$e(\delta; m_0) \coloneqq \hat{\phi}_{m_0}(\delta)  - \phi(m_0 + \delta)$. Two
examples are shown in \autoref{fig:approx-error}. For an instance right at
the decision boundary, improving the classification margin to exceed one
causes the quadratic approximation to over-estimate the true loss value, leading to
smaller proposed update vectors $\vect{p}$. On the other hand, for any instance
classified with at least margin $1$, the quadratic approximation cannot "see"
the increased error as the margin shrinks, and may propose overly large steps
$\vect{p}$ that have to be shrunk using the line search. The practical
consequences of this can be seen in the next section.

\section{Analysis of Criteria for Initial Weights}
\label{sec:analysis}


\paragraph{Low Loss as a Criterion for Initial Weights}
In \textcite{fang2019fast}, the authors showed (under some conditions, see
their Lemma 3.1) that the optimal weight vector $\weightvec^*(\labels)$ for
one labeling $\labels$ results in good performance also for a different
labeling $\labels^\prime$ if the two label vectors are close in Hamming distance
$\lham$:
\begin{equation}
    \mathcal{L}[\weightvec^*(\labels), \labels^\prime] - \mathcal{L}[\weightvec^*(\labels^\prime), \labels^\prime] \leq \text{const} \cdot \lham(\labels, \labels^\prime).
\end{equation}
Given that for tail labels we have $\lham(\labels, \vect{-1}) \ll
\numinstances$, they conclude that using $\weightvec^*(-\vect{1})$ makes the
initial vector result in much lower loss than using zero-initialized weights.
Then, based on convergence rates of iterative minimizers, they estimate the
speedup that $\init=\weightvec^*(-\vect{1})$ provides over $\init=\vect{0}$.

However, this considers only one of the three criteria for fast convergence we
listed in the introduction. As we show below, it is entirely possible for an
initial vector to induce an initial loss that is orders of magnitude smaller,
but still not provide any speedup. The reason is that their estimation is
based on asymptotic convergence rates based on the number of iterations. This does not
need to agree with actual computation time, for example if the computations
become faster closer to the minimum due to implicit negative mining as
discussed above. 

Furthermore, certain regions in the weight space may be more benign towards
the chosen minimization procedure than others. For example, in deep networks
one often chooses an initialization procedure that preserves variance and mean
over layers, to prevent vanishing or exploding gradients
\cite{glorot2010understanding}. Such a technique does not provide any reduced
loss for the starting point, but is very effective in speeding up the
training. For the concrete case of Newton optimization, this means that we
require the local quadratic approximation to fit the true loss well, which may
become problematic if many points are classified with margin
close to 1.

The reason why the method of \textcite{fang2019fast} works well, even though
its derivation only considered the reduction in loss, is that as a byproduct
it also increases sparsity, and appears to produce vectors that are amenable
to Newton optimization. Below, we construct an initialization method that also
induces a strong reduction in loss, but does not provide sparsity or useful
second-order approximations.

\paragraph{Running Example}
To evaluate the initialization methods and investigate their properties, we
take the \texttt{AmazonCat-13K} dataset from the extreme classification
repository \cite{Bhatia16,mcauley2013hidden} as a running example in this
section and the next. The input features are sparse tf-idf values of a
bag-of-words representation augmented by an additional bias feature that is
set to 1. The train/test split is taken from the repository.

\paragraph{Bias-Initialization}
A simple way to calculate an approximation to $\weightvec^*(-\vect{1})$, the
optimal weight-vector that predicts the absence of the label for every
instance, is to use a weight vector $\weightvec_{\mathrm{b}} = (-1,
0,\ldots)\transposed$, where we assume that the bias feature is at index
$0$.\footnote{Such a strategy seems to have been considered by
\textcite{fang2019fast}, as it can be found in their code at
\url{https://github.com/fanghgit/XMC}, though it is not mentioned in the
paper.}

\begin{figure}
    \input{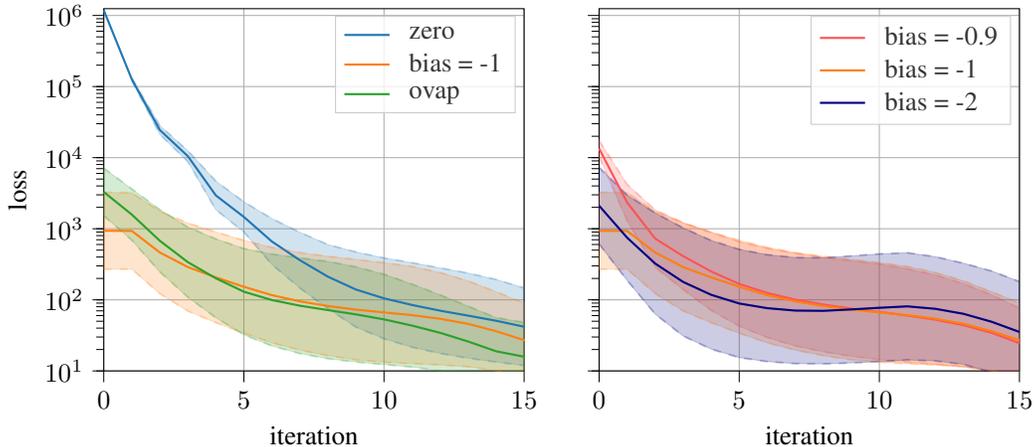}
    \caption{Objective value at different iterations for starting with \texttt{zero}
    vector, \texttt{ovap} and with \texttt{bias}-initialization
    $\weightvec_{b}$. The right side shows the progress for different values
    of the bias weight. The shaded area spans the 5\% to 95\% quantiles. Note
    that the slight increase in objective for the later iterations is just an
    artifact due to the fact that we plot the loss values averaged for the binary sub-problems that have not yet terminated at 
    the given iteration. The sub-problems with low loss value
    terminate earlier, causing an increase in average of the remaining trajectories.}
    \label{fig:bias-objective}
\end{figure}

Then, the score for any instance $\instance$ will be $m=-1$, which means that
it is classified as negative with margin one, so this weight vector is a
minimizer of the squared hinge loss (without regularization). As can be seen
in \autoref{fig:bias-objective} (left), the initial value of the loss is decreased by
three orders of magnitude, in fact to a lower value than with the
\texttt{ovap} initializer. Yet after some iterations, the ovap based
optimization overtakes \texttt{bias}. In particular, the
\texttt{bias}-initialized optimization does not make any substantial progress
during the first iteration.

This can be explained by the fact that by choosing $-1$ for the bias weight,
the initial weights are exactly at a position where there is a discontinuity
in the Hessian. This suggests that it might be beneficial to use a different
value for the bias, \eg $\init = 0.9\weightvec_{\mathrm{b}}$ or $\init = 2
\weightvec_{\mathrm{b}}$. This intuition is confirmed by \autoref{fig:bias-objective}
(right), where we can see that both variations make significant progress in
the first iteration, with $2\weightvec_{\mathrm{b}}$ clearly outperforming
$\weightvec_{\mathrm{b}}$.

So far, we only have looked at the optimization time in terms of number of
iterations, but what we really care about is wall-clock time. Here, the
difference between the three \texttt{bias}-init variations becomes even more
pronounced. For $2\weightvec_{\mathrm{b}}$ we get $558$ seconds, much faster than
$\weightvec_{\mathrm{b}}$ and $0.9\weightvec_{\mathrm{b}}$ at $1161$ and $995$ seconds. This is
because the calculations of Hessian-vector products become sparse, as only
those instances with non-zero loss need to be taken into account. For $\lambda
\weightvec_{\mathrm{b}}$, $\lambda\geq 1$, these are all the negatives $\negatives$.
However, for $\lambda=1$ the very first optimization step destroys the
sparsity, so even though the first iteration is fast, it does not make
progress and makes the following iterations slow again. To a much lesser
degree, this also happens for $\lambda=2$ as can be seen from
\autoref{fig:iter-dur-sparse}. For $\lambda < 1$, we are not at the
discontinuity of the Hessian, but the quadratic approximation now pessimized
the actual loss function, and there is no sparsity in the first iteration.

\paragraph{OvA-Primal}
When running the experiments with the OvA-primal initial vector, we noticed
that the choice of stopping criterion for learning the initial vector can have
a strong influence on performance. In particular, if we use a criterion as
strict as for the actual binary problems, the training time is much larger
than for a loose stopping criterion. This can be explained by the fact that
the OvA-primal task will converge to a solution with similar characteristics
as $\weightvec_{b}$, \ie each training instance will be close to the decision
boundary. For the main paper, we have used a stopping criterion of $\|\nabla
\mathcal{L}[\vect{w}^*]\| \leq 0.01 \|\nabla \mathcal{L}[\vect{0}]\|$.  

A more detailed discussion of the training dynamics with bias and OvA-primal
initialization can be found in the supplementary.

\section{Average-of-Positives Initialization}
\label{sec:aop-init}
\begin{figure}
    \centering
    \input{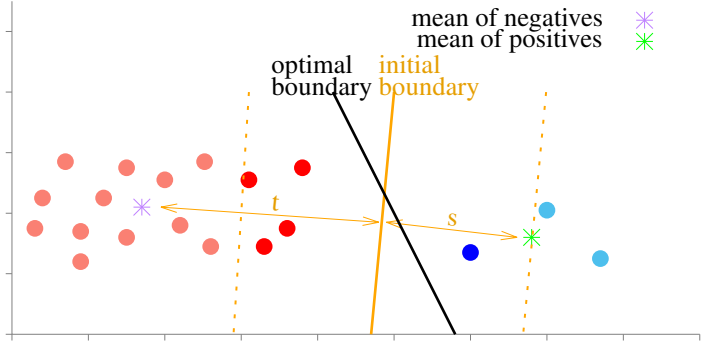}
    \caption{Motivation for the Average-of-Positives initialization. If the
    data is linearly separable, then the feature-means of the positives (blue)
    and the negatives (red) will lie on the correct side of the optimal separating
    hyperplane (black line). Thus we may restrict our search for the initial
    vector to the set of the separating planes of $\posmean$ and $\negmean$ (orange), which
    have to lie in the plane spanned by $\posmean$ and $\negmean$ if we want a
    minimum-norm solution. In order to achieve strong implicit negative mining
    in the first epoch, we choose a solution where the center of mass of the
    negatives is classified correctly with a large margin (dashed orange lines), so that most of the
    negatives (light red) will be classified correctly with a margin over more than 1 and thus not enter the Hessian computation.}
     \label{fig:avg-of-pos-sketch}
\end{figure}

\paragraph{Motivation}
In this section, we derive a simple way to generate an initial vector which is
motivated by the following observation: If the data is linearly separable with
margin, then the final weight vector will separate the convex hull of the
negatives $\negatives$ from the positives $\positives$, which in particular
implies that it separates the centres of mass $\posmean$ of the positives and
$\negmean$ of the negatives. This means that we can restrict the search space
of weight vectors to those that separate $\posmean$ and $\negmean$. A sketch
of this situation is given in
\autoref{fig:avg-of-pos-sketch}.

\paragraph{Derivation}
Without making use of any additional information about the training data (as
we want a computationally cheap procedure), there are only very general
conditions we can impose to choose among these hyperplanes. As a first step, we
can parameterize the search space, based on the margins of $\posmean$ and
$\negmean$, that is we impose
\begin{equation}
    \langle \init, \posmean \rangle = s, \quad
    \langle \init, \negmean \rangle = t,
\end{equation}
for two hyperparameters $s$ and $t$.

These are only two linear constraints in the high dimensional weight space.
However, for good generalization, we prefer weight vectors with minimal norm.
If we want a minimum $L_2$-norm solution, then the search space becomes
restricted to $\operatorname{span}(\posmean, \negmean)$, and we can
parameterize $\init = u \posmean + \tilde{v} \negmean$. This leads to a system of two
linear equations in the two unknowns $u$ and $v$, with a unique solution
except in the unlikely event that $\posmean$ and $\negmean$ are linearly
dependent.

\paragraph{Reparameterization for Efficiency}
Since most labels have only few positive instances, their mean $\posmean$ can
be calculated quickly, but calculating $\negmean$ directly would be an
$O(\numinstances)$ operation. However, we can precompute the mean of all instances
$\ftrmean$, and use the property
\begin{equation}
    |\negatives| \negmean + |\positives| \posmean = |\dataset| \ftrmean.
\end{equation}
Because $\operatorname{span}(\posmean, \negmean) =
\operatorname{span}(\posmean, \ftrmean)$, we can make the equivalent
parameterization $\init = u \posmean + v \negmean$, leading to the equations
\begin{align}
     s = \langle \init, \posmean \rangle &= \langle u \posmean + v \ftrmean, \posmean\rangle = v \langle \ftrmean, \posmean\rangle + u \langle \posmean, \posmean\rangle\\
     |\negatives| t = |\negatives|  \langle \init, \negmean \rangle &= \langle u \posmean + v \ftrmean, |\dataset| \ftrmean - |\positives| \posmean \rangle.
\end{align}
Defining $\alpha \coloneqq |\positives| / |\dataset|$ leads to
\begin{align}
     u &= \frac{\langle \ftrmean, \posmean\rangle (t + (s - t) \alpha ) - s \langle \ftrmean, \ftrmean \rangle}
{\langle \posmean,  \ftrmean \rangle^2 - \langle \posmean, \posmean\rangle \langle \ftrmean, \ftrmean \rangle} \label{eq:calc-u} \\
v &= (s - u \langle \posmean, \posmean\rangle) / \langle \ftrmean, \posmean\rangle. \label{eq:calc-v}
\end{align}
A short discussion on the situations if the denominators become zero is given
in the supplementary.

As we want the loss vector corresponding to the initial weights to be sparse,
we want the distance of $\negmean$ to the decision boundary to be larger
than that of $\posmean$. Empirically, we found $s=1$, $t=-2$
(cf. supplementary) to work well.

\begin{figure}
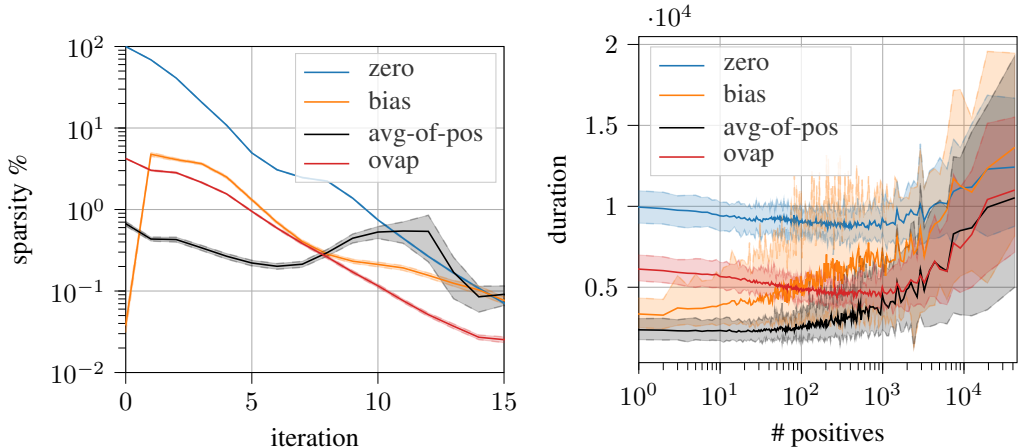

    \input{figures/IterSparsity}
    \input{figures/Duration}
    \caption{On the left, the sparsity (in percent of nonzeros) of the Hessian
    calculation over different training epochs is shown. Even though
    bias-initialization starts out very sparse, this property is lost after
    the first update step. On the right, the training time for a single
    label is shown, depending on the number of positives for that label. The shaded
    area marks the 5\% to 95\% quantile.}
    \label{fig:iter-dur-sparse}
\end{figure}

\paragraph{Evaluation}
Does this initialization method work as we expect? As shown in
\autoref{fig:iter-dur-sparse}, this method induces loss vectors that
are a bit more sparse than the ones for \texttt{ovap}, and keep this property
over the course of training, inducing faster update times for each iteration.
As expected, we get the most benefit for tail labels, where the assumptions of \autoref{fig:avg-of-pos-sketch} are much better fulfilled than for head labels. Still, our method is faster or at least as fast as any of the other methods across the entire range of number of positives. Additional graphs showing the number of iterations, duration per iteration etc. can be found in the supplementary.

\section{Experimental Evaluation}
\label{sec:experiments}

\paragraph{Infrastructure and Training Configuration}
Our code is a re-implementation of the conjugate-gradient Newton procedure of
recent versions of Liblinear \cite{galli2021study}. The hyperparameters
related to stopping condition for conjugate-gradient iterations ($\epsilon =
0.5$) preconditioner ($\alpha=0.01$), back-tracking line search ($\alpha=0.5$,
$\eta=0.01$, $\text{max\_steps}=20$), and stopping condition for the
optimization ($\epsilon=0.01$) have been taken from their code. To reduce
model size, all weights below a threshold of $0.01$ have been clipped to zero.

All our experiments were run on a single compute node equipped with two AMD
Rome 7H12 CPUs. The code is NUMA-aware and produces a copy of the training
data for each NUMA-node, but memory bandwidth still is the limiting factor in
scalability. We observed very good scaling for $64$ cores, but sublinear
improvements for more cores. This means that the reported times here cannot be
directly converted to single-core timings by multiplying with $128$.

\paragraph{Sparse Data}
We ran tests using the \texttt{Eurlex-3k} \cite{mencia2008efficient},
\texttt{AmazonCat-13k} \cite{mcauley2013hidden},
\texttt{WikiSeeAlsoTitles-350K}, \texttt{WikiTitles-500k},
\texttt{Amazon-670k} \cite{mcauley2013hidden}, \texttt{WikiLSHTC-325k}
\cite{partalas2015lshtc}, \texttt{Amazon-3M}
\cite{mcauley2015image,mcauley2015inferring}, and \texttt{Wikipedia-500k}
\cite{Bhatia16} in their train/test splits as available from
\textcite{Bhatia16}. For the latter dataset, we could not train the model
within the 36 hour job time limit on our cluster system, even with the
improved initialization. For \texttt{Amazon-3M} we have only run our
initializer due to the large computational cost involved, so we cannot report
the speed-up, but we have demonstrated that learning a DiSMEC-style model is
feasible even for datasets with more than a million labels.

As shown in \autoref{tabel:timings}, our method is 3 to 4 times faster than
\texttt{zero} initialization, and also significantly faster than \texttt{bias}
and \texttt{ovap} initialization. Based on the timing reported in
\textcite{fang2019fast}, we have also calculated the speedup of the
\texttt{OvA-Primal++} method introduced there, for the datasets for which they
have run tests. 

Because the optimization problem is convex, the minimum is not affected by
the choice of initial parameter. However, because in practice only an
approximate minimum is found, there are slight variations (around 0.1\%) in
precision@k metrics for the different methods. The concrete numbers can be
found in the supplementary.

\paragraph{Dense Data}
We also ran the tests on the same dataset of dense features as used for slice
\footnote{\url{http://manikvarma.org/code/Slice/download.html}}. We noticed
that the default setting for the stopping condition as specified by Liblinear,
$\epsilon=0.01$, is far too strict in this setting. Therefore, we've increased
this parameter for $\epsilon=1$ here. For these settings, we observed larger
variability in the classification accuracy.

\begin{table}
\centering
\caption{Training time for different datasets and initialization methods. \textit{The
timings are given in minutes}. For the logistic loss, we have changed to $t =
-3$. Ratio denotes the time taken with \texttt{zero}-initialization divided by
\texttt{aop}-initialization, the \texttt{ovap++} column is the relative
speedup of the spanning-tree based OvA-Primal++ method, calculated from the
values reported in \textcite{fang2019fast}.}
    \begin{filecontents*}{timings.txt}
Dataset             Setting     zero    bias    aop     ovap        ratio       ovap++
Eurlex-4k           tf-idf      11      8       6       7           1.8         1.9
Amazoncat-13k       tf-idf      1049    557     348     640         3.0         1.8
Amazoncat-14k       tf-idf      4563    1790    1217    2497        3.7         2.1
Wiki10-31k          tf-idf      239     215     141     189         1.7         1.8
WikiLSHTC           tf-idf      66660   28769   18628   41278       3.6         {}
WikiTitles-500k     tf-idf      17559   5598    4310    18920       4.1         {}
Amazon-670k         tf-idf      19026   8926    5045    11840       3.8         {}
Amazon-3M           tf-idf      {}      {}      69774   {}          {}          {}
Eurlex-4k           slice       8       4       3       4           2.7         {}
Amazon-670k         slice       46945   22589   19272   29350       2.4         {}
Amazoncat-13k       logistic    1758    1517    1410    1214        1.3         {}
\end{filecontents*}
\pgfplotstabletypeset[
    every head row/.style={before row={\toprule}, after row=\midrule,},
    every last row/.style={after row=\bottomrule}, 
    columns/Dataset/.style={string type, column type=l},
    columns/Setting/.style={string type, column type=l},
    columns/zero/.style={divide by=60, fixed, fixed zerofill, precision=2, column type=r},
    columns/bias/.style={divide by=60, fixed, fixed zerofill, precision=2, column type=r},
    columns/aop/.style={divide by=60, fixed, fixed zerofill, precision=2, column type=r},
    columns/ovap/.style={divide by=60, fixed, fixed zerofill,  precision=2, column type=r},
    columns/ratio/.style={fixed zerofill, precision=1, column name={Ratio}},
    columns/ovap++/.style={fixed zerofill, precision=1},
    empty cells with={\textemdash}
    ]
    {timings.txt}
    \label{tabel:timings}
\end{table}

\paragraph{Logistic Loss}
We have also run a test where we replaced the squared hinge loss with the
logistic loss. As the logistic loss only vanishes asymptotically, this is a
setting where we cannot benefit from implicit negative mining. This means that
later iterations will take approximately as long as earlier ones, and there is
much less benefit from being able to skip the first iterations. The result is
a much reduced benefit from our proposed initialization. 

Due to the vast number of negatives (for a tail label), the loss function's
minimum will not be achieved when the negatives are classified correctly with
margin one as is the case with squared-hinge, but will in fact train a larger
margin for the negatives. This suggests that by decreasing the $t$ parameter
further, our initial guess for the separating hyperplane will be closer to the
truth.

\section{Discussion}
We have provided a way to initialize the weights of a linear OvA extreme
classifier in such a way as to reduce training times. Our experiments show
that aside from the initial loss value investigated in
\textcite{fang2019fast}, the implicit sparsity and local smoothness properties
of the loss landscape also play an important role in the success of the
method.

\paragraph{Outlook}
The choice of initial vector is an underexplored design tool in XMC that is
orthogonal to many other design choices such as explicit negative mining \cite{reddi2019stochastic,jain2019slice},
training meta classifiers over buckets of labels \cite{medini2019extreme,dahiya2021deepxml}, or the choice of
regularizer. Future work should thus look into combining these, for example
integrating our initialization into the OvA parts of Slice\cite{jain2019slice}, Parabel \cite{prabhu2018parabel}, Probabilistic Label Trees \cite{Wydmuch_et_al_2018}, 
or Astec \cite{dahiya2021deepxml}.

\paragraph{Limitations}
The initialization method discussed in this work is mainly applicable to linear one-vs-rest XMC algorithms. This rules out label-embedding schemes \cite{guo2019breaking,bhatia2015sparse}, decision-tree based classifiers \cite{prabhu2014fastxml, majzoubi2020ldsm} and deep-learning methods in which the classifier is jointly learnt with the intermediate representations  \cite{NEURIPS2019_9e6a921f}.


\begin{ack}
We wish to acknowledge CSC – IT Center for Science, Finland, as well as the
Aalto Science-IT project, for providing the required computational resources.
We would also like to thank Mohammadreza Qaraei for his help in preparing
\autoref{fig:avg-of-pos-sketch}.
\end{ack}

\printbibliography

\clearpage
\section{Supplementary}
\subsection{What happens if the denominator of \texttt{aop} is zero}
\FloatBarrier
The expression in equation \eqref{eq:calc-u} diverges if $\posmean$ and
$\ftrmean$ are linearly dependent, which implies that the means of features
for negatives and positives only differs by a factor. In that case, we cannot
(except in unlikely special cases) find an initial vector that fulfills our
condition, so we return the zero vector.

A second numerical instability occurs if $\abs{\langle \ftrmean, \posmean
\rangle} \ll 1$, during the calculation of $v$. To investigate this setting,
we substitute $\langle \ftrmean, \posmean \rangle$ with $0$ in
\eqref{eq:calc-u} and \eqref{eq:calc-v}.
\begin{align}
     s &= u \langle \posmean, \posmean\rangle\\
     |\negatives| t &= v |\dataset|  \langle \ftrmean, \ftrmean \rangle  - u |\positives| \langle \posmean,  \posmean \rangle
\end{align}
This leads to
\begin{equation}
u = \frac{s}{\langle \posmean, \posmean\rangle}, \qquad
v = \frac{|\negatives| t + |\positives| s}{|\dataset| \langle \ftrmean, \ftrmean \rangle}.
\end{equation}
\subsection{Additional graphs for training with \texttt{bias} initialization}
\begin{figure}[H]
    \centering
    \input{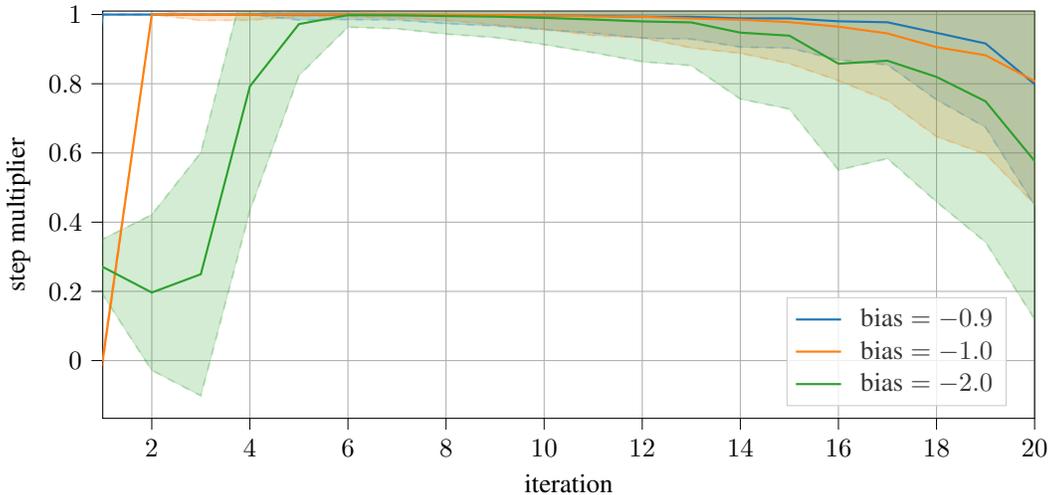}
    \caption[Multiplier found by backtracking line search]{
    Multiplier found by backtracking line search, \ie the fraction of the
    descent direction as found by the CG solver that can be applied to the
    real objective function and still lead to sufficient reduction in loss.
    The shaded area indicates $\pm$ one standard deviation. Having a value
    smaller than one indicates that the quadratic approximation to the loss is
    overly optimistic in the given direction, and the real step has to be
    scaled down. As discussed in the main text (\autoref{fig:approx-error}),
    this happens when the direction in question is such that it reduces the
    margin of an instance that is classified correctly with margin more than
    one. This effect can be seen in the graph here:

    For $\text{bias} = -0.9$, all the negative instances still have nonzero
    error, so the update will choose a direction that makes them more
    negative, which is a direction in which the quadratic approximation is
    overestimating the loss.

    On the other hand, with $\text{bias} = -1$, the only nonzero signal is
    coming from the positive instances, which will cause the bias to be
    reduced. Thus, most of the negative instances will get nonzero error, and
    the quadratic approximation severly underestimates the true loss.
    Consequently, the line search shows that only a minute step towards the
    desired direction is allowed.

    For $\text{bias} = -2$, the situation is similar, but this time the
    distance for each negative instance until it gets nonzero error is much
    larger, and as such the underestimation effect is reduced and larger
    stepsize multipliers are possible.}
\end{figure}

\begin{figure}[H]
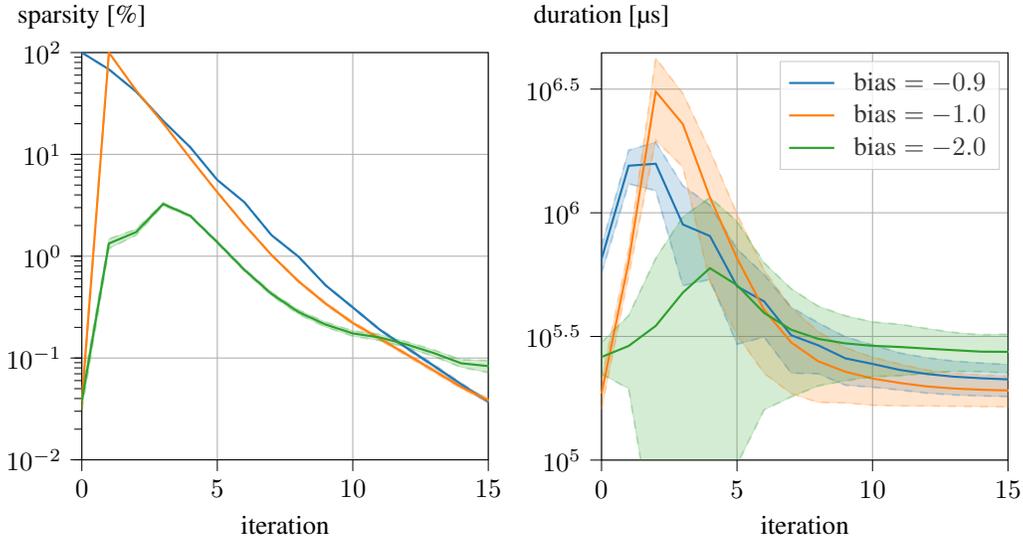

\centering
\input{supplementarty/bias/IterSparsity}%
\input{supplementarty/bias/IterDuration}
\caption[Sparsity and Iteration Duration]{
The average sparsity (in percent of nonzeros) and corresponding duration (in
µs) of one iteration of the Newton optimization. The averages are taken over
the individual binary problems, and the shaded area shows the $2\sigma$ error
of the mean. The total computation time includes a (sparse) matrix
multiplication and a varying number of CG steps. As this data shows, even
though $-1$ and $-2$ start very sparse, much of the sparsity is lost during
the first training steps. For the later iterations, $-2$ results in slightly
less sparsity and slower iterations, however the effect on total running time
is far overshadowed by the much faster earlier iterations (note the
logarithmic axes). As \autoref{fig:bias:iterations} shows that
$\text{bias}=-2$ finishes after at most 15 steps, this graph also indicates
that the approximate minima found by the different initialization procedures
have slightly different characteristics: The larger sparsity in $-0.9$ and
$-1.0$ setting means that fewer instances need to have a larger contribution
to the overall error.}
\end{figure}

\begin{figure}[H]
    \input{supplementarty/bias/Iterations}
    \caption{Number of Newton optimization steps required for convergence. On the left,
    the data is shown as a histogram, with the bins corresponding to the
    number of binary problems that required the corresponding number of steps.
    On the right, the average number of steps, and the 95\% quantiles, are
    plotted as a function of the number of positive instances. The data shows
    that the benefit of the $\text{bias}=-2$ initialization is not limited
    purely to faster iterations because of increased sparsity, but it also
    needs fewer steps, despite having slightly larger initial loss than
    $\text{bias}=-1$. This indicates that each step has to make more progress,
    \ie the loss landscape is more benign to the Newton optimization around
    the $\text{bias}=-2$ trajectory. The right-hand graph also shows that most
    of the benefit comes from the tail labels.}
    \label{fig:bias:iterations}
\end{figure}

\begin{figure}[H]
    \input{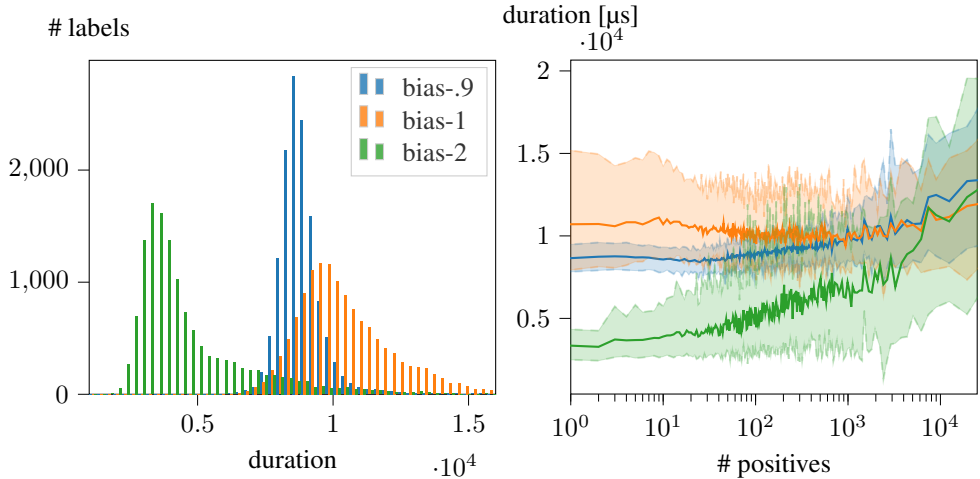}
    \caption{Total duration of the individual binary problems. The histogram
    on the left has bins corresponding to the number of binary problems taking
    a certain amount of time, on the right the average time and 95\% quantiles
    are shown in dependence of the number of positive instances. Duration is
    measured in milliseconds. The graphs show that $\text{bias} = -1$} is a
    local maximum in terms of computation time.
\end{figure}

\subsection{Additional graphs for training with \texttt{ovap} initialization}
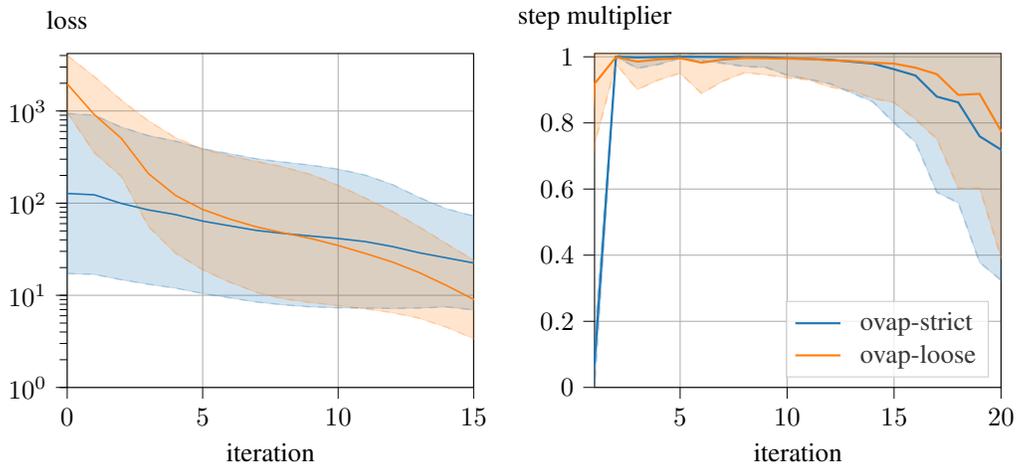
\begin{figure}[H]
\begin{tikzpicture}

\definecolor{color0}{rgb}{0.12156862745098,0.466666666666667,0.705882352941177}
\definecolor{color1}{rgb}{1,0.498039215686275,0.0549019607843137}

\begin{axis}[
log basis y={10},
tick align=outside,
tick pos=left,
x grid style={white!69.0196078431373!black},
xlabel={iteration},
xmajorgrids,
xmin=0, xmax=15,
xtick style={color=black},
y grid style={white!69.0196078431373!black},
ymajorgrids,
ymin=1, ymax=4204.68746914682,
ymode=log,
ytick style={color=black},
every axis y label/.style={at={(current axis.north west)},above=2mm},
ylabel={loss},
width=0.5\linewidth
]
\path [draw=color0, fill=color0, opacity=0.2]
(axis cs:0,17.2181648691785)
--(axis cs:0,942.029435449955)
--(axis cs:1,901.237050771694)
--(axis cs:2,667.997826136783)
--(axis cs:3,540.605894934081)
--(axis cs:4,471.820007492084)
--(axis cs:5,390.073114362998)
--(axis cs:6,342.83473458863)
--(axis cs:7,302.494831369954)
--(axis cs:8,278.938659761831)
--(axis cs:9,258.592278180041)
--(axis cs:10,233.73151273934)
--(axis cs:11,201.844198609792)
--(axis cs:12,159.271942986405)
--(axis cs:13,114.967290281335)
--(axis cs:14,86.8531870854708)
--(axis cs:15,72.5476153294341)
--(axis cs:16,64.1601401492003)
--(axis cs:17,71.222535916962)
--(axis cs:18,84.6889282525613)
--(axis cs:19,86.7006742357603)
--(axis cs:20,94.1875729062113)
--(axis cs:21,80.54090529268)
--(axis cs:22,83.7099593860291)
--(axis cs:23,62.3734163896631)
--(axis cs:24,60.4807223969405)
--(axis cs:25,65.5583015396225)
--(axis cs:26,19.4884226809593)
--(axis cs:27,12.9552285143297)
--(axis cs:28,16.2372237531141)
--(axis cs:29,14.5176960690815)
--(axis cs:29,14.5058144378476)
--(axis cs:29,14.5058144378476)
--(axis cs:28,5.34481735960331)
--(axis cs:27,4.99061717869242)
--(axis cs:26,5.18448407214524)
--(axis cs:25,6.98776991783268)
--(axis cs:24,7.9592246316003)
--(axis cs:23,8.74444168178498)
--(axis cs:22,9.18851604562117)
--(axis cs:21,8.44086241873903)
--(axis cs:20,8.94220475055388)
--(axis cs:19,8.92055269029468)
--(axis cs:18,7.78268047045736)
--(axis cs:17,6.92027595482957)
--(axis cs:16,6.38810720929185)
--(axis cs:15,6.93653984142152)
--(axis cs:14,7.47923651151233)
--(axis cs:13,7.27025738247937)
--(axis cs:12,7.19238144724237)
--(axis cs:11,7.25987515505802)
--(axis cs:10,7.31954424201511)
--(axis cs:9,7.48213289518988)
--(axis cs:8,7.8660408674346)
--(axis cs:7,8.43629967758198)
--(axis cs:6,9.39671189488267)
--(axis cs:5,10.444475452724)
--(axis cs:4,11.9554256317768)
--(axis cs:3,13.1708701643175)
--(axis cs:2,14.7452090731749)
--(axis cs:1,16.8476609680243)
--(axis cs:0,17.2181648691785)
--cycle;

\path [draw=color1, fill=color1, opacity=0.2]
(axis cs:0,970.36710255075)
--(axis cs:0,4004.57352686758)
--(axis cs:1,2368.25238802887)
--(axis cs:2,1310.45749422127)
--(axis cs:3,782.684620418072)
--(axis cs:4,513.302846257653)
--(axis cs:5,385.705758471949)
--(axis cs:6,325.401517813911)
--(axis cs:7,283.38783844731)
--(axis cs:8,246.391379731528)
--(axis cs:9,204.464814651073)
--(axis cs:10,156.118895528999)
--(axis cs:11,113.780685236123)
--(axis cs:12,81.4483707310741)
--(axis cs:13,55.1182111209645)
--(axis cs:14,36.2946623145295)
--(axis cs:15,24.1165509456247)
--(axis cs:16,17.9085417836315)
--(axis cs:17,17.3806836083496)
--(axis cs:18,21.6677155258826)
--(axis cs:19,29.2048871207322)
--(axis cs:20,40.9032037058468)
--(axis cs:21,44.436403388896)
--(axis cs:22,70.7614863904805)
--(axis cs:23,72.5898799650254)
--(axis cs:24,73.519064699982)
--(axis cs:25,50.7022567967918)
--(axis cs:26,56.990526827517)
--(axis cs:27,84.8232741983839)
--(axis cs:28,113.85080870872)
--(axis cs:28,13.2112740243399)
--(axis cs:28,13.2112740243399)
--(axis cs:27,12.3964387929171)
--(axis cs:26,5.37705603859469)
--(axis cs:25,4.22056670522727)
--(axis cs:24,5.00520456090684)
--(axis cs:23,6.07950850253821)
--(axis cs:22,4.34765167681553)
--(axis cs:21,3.71350158107511)
--(axis cs:20,3.69128165068137)
--(axis cs:19,3.09665300843555)
--(axis cs:18,2.41863510836862)
--(axis cs:17,2.29468128280437)
--(axis cs:16,2.59698232820889)
--(axis cs:15,3.3869183666003)
--(axis cs:14,4.50392850460114)
--(axis cs:13,5.63234119957792)
--(axis cs:12,6.48492966256363)
--(axis cs:11,7.11994943545514)
--(axis cs:10,7.72993586689389)
--(axis cs:9,8.33387403858657)
--(axis cs:8,9.13587397198438)
--(axis cs:7,10.7353532589289)
--(axis cs:6,13.8669839384977)
--(axis cs:5,18.920091014447)
--(axis cs:4,28.607595938192)
--(axis cs:3,55.861330430107)
--(axis cs:2,194.634916022267)
--(axis cs:1,352.518484935902)
--(axis cs:0,970.36710255075)
--cycle;

\addplot [semithick, color0]
table {%
0 127.357834981584
1 123.222304325246
2 99.2459954195293
3 84.3815741275394
4 75.1053194597826
5 63.828747972461
6 56.7584286999554
7 50.5167006875603
8 46.8416790603625
9 43.9866092238647
10 41.361916641383
11 38.280069000458
12 33.8458944009708
13 28.9109285722808
14 25.4871639848545
15 22.4327756671514
16 20.2450451675535
17 22.2008919358691
18 25.6730767142385
19 27.8103925323641
20 29.0214500307111
21 26.0736399579384
22 27.7339197553614
23 23.3542437709531
24 21.9403658911767
25 21.4034185905611
26 10.0517370131071
27 8.04080754511011
28 9.31584646650038
29 14.5117540374403
};
\addplot [semithick, color0, opacity=0.2, dashed, forget plot]
table {%
0 942.029435449955
1 901.237050771694
2 667.997826136783
3 540.605894934081
4 471.820007492084
5 390.073114362998
6 342.83473458863
7 302.494831369954
8 278.938659761831
9 258.592278180041
10 233.73151273934
11 201.844198609792
12 159.271942986405
13 114.967290281335
14 86.8531870854708
15 72.5476153294341
16 64.1601401492003
17 71.222535916962
18 84.6889282525613
19 86.7006742357603
20 94.1875729062113
21 80.54090529268
22 83.7099593860291
23 62.3734163896631
24 60.4807223969405
25 65.5583015396225
26 19.4884226809593
27 12.9552285143297
28 16.2372237531141
29 14.5176960690815
};
\addplot [semithick, color0, opacity=0.2, dashed, forget plot]
table {%
0 17.2181648691785
1 16.8476609680243
2 14.7452090731749
3 13.1708701643175
4 11.9554256317768
5 10.444475452724
6 9.39671189488267
7 8.43629967758198
8 7.8660408674346
9 7.48213289518988
10 7.31954424201511
11 7.25987515505802
12 7.19238144724237
13 7.27025738247937
14 7.47923651151233
15 6.93653984142152
16 6.38810720929185
17 6.92027595482957
18 7.78268047045736
19 8.92055269029468
20 8.94220475055388
21 8.44086241873903
22 9.18851604562117
23 8.74444168178498
24 7.9592246316003
25 6.98776991783268
26 5.18448407214524
27 4.99061717869242
28 5.34481735960331
29 14.5058144378476
};
\addplot [semithick, color1]
table {%
0 1971.27025296328
1 913.702765550027
2 505.035428795354
3 209.097594925759
4 121.179042823678
5 85.425921447097
6 67.1739355783796
7 55.1567634566799
8 47.4447109065973
9 41.2793412087086
10 34.7388694411529
11 28.4624792597665
12 22.9823183191214
13 17.6194373163175
14 12.7854825549637
15 9.03774249117488
16 6.81968962168232
17 6.31530912611762
18 7.23921939776958
19 9.50985812530533
20 12.2876053522838
21 12.8458029815951
22 17.5398487724248
23 21.0073985168464
24 19.1827515740041
25 14.6284741827855
26 17.5054636162712
27 32.4269413731096
28 38.7829115944587
29 30.6554163344165
30 30.6459236462377
};
\addplot [semithick, color1, opacity=0.2, dashed, forget plot]
table {%
0 4004.57352686758
1 2368.25238802887
2 1310.45749422127
3 782.684620418072
4 513.302846257653
5 385.705758471949
6 325.401517813911
7 283.38783844731
8 246.391379731528
9 204.464814651073
10 156.118895528999
11 113.780685236123
12 81.4483707310741
13 55.1182111209645
14 36.2946623145295
15 24.1165509456247
16 17.9085417836315
17 17.3806836083496
18 21.6677155258826
19 29.2048871207322
20 40.9032037058468
21 44.436403388896
22 70.7614863904805
23 72.5898799650254
24 73.519064699982
25 50.7022567967918
26 56.990526827517
27 84.8232741983839
28 113.85080870872
29 nan
30 nan
};
\addplot [semithick, color1, opacity=0.2, dashed, forget plot]
table {%
0 970.36710255075
1 352.518484935902
2 194.634916022267
3 55.861330430107
4 28.607595938192
5 18.920091014447
6 13.8669839384977
7 10.7353532589289
8 9.13587397198438
9 8.33387403858657
10 7.72993586689389
11 7.11994943545514
12 6.48492966256363
13 5.63234119957792
14 4.50392850460114
15 3.3869183666003
16 2.59698232820889
17 2.29468128280437
18 2.41863510836862
19 3.09665300843555
20 3.69128165068137
21 3.71350158107511
22 4.34765167681553
23 6.07950850253821
24 5.00520456090684
25 4.22056670522727
26 5.37705603859469
27 12.3964387929171
28 13.2112740243399
29 nan
30 nan
};
\end{axis}
\end{tikzpicture}
\begin{tikzpicture}

\definecolor{color0}{rgb}{0.12156862745098,0.466666666666667,0.705882352941177}
\definecolor{color1}{rgb}{1,0.498039215686275,0.0549019607843137}

\begin{axis}[
legend cell align={left},
legend style={fill opacity=0.8, draw opacity=1, text opacity=1, at={(0.03,0.97)}, anchor=north west, draw=white!80!black},
legend pos={south east},
tick align=outside,
tick pos=left,
x grid style={white!69.0196078431373!black},
xmin=1, xmax=20,
xlabel={iteration},
xmajorgrids,
ymajorgrids,
xtick style={color=black},
y grid style={white!69.0196078431373!black},
ymin=0.0, ymax=1.01,
ytick style={color=black},
width=0.5\linewidth,
every axis y label/.style={at={(current axis.north west)},above=2mm},
ylabel={step multiplier},
]
\path [draw=color0, fill=color0, opacity=0.2]
(axis cs:0,0)
--(axis cs:0,0)
--(axis cs:1,0.149900638515719)
--(axis cs:2,1)
--(axis cs:3,1.03106416109964)
--(axis cs:4,1.02133601477806)
--(axis cs:5,1.00429315618266)
--(axis cs:6,1.0114147401939)
--(axis cs:7,1.01843728811957)
--(axis cs:8,1.02660945826298)
--(axis cs:9,1.02864024252864)
--(axis cs:10,1.04691049621491)
--(axis cs:11,1.05511643322051)
--(axis cs:12,1.06348176524353)
--(axis cs:13,1.07932896056438)
--(axis cs:14,1.09493012717027)
--(axis cs:15,1.12417689967966)
--(axis cs:16,1.1452337575592)
--(axis cs:17,1.16881955737926)
--(axis cs:18,1.1659172640736)
--(axis cs:19,1.14102919953125)
--(axis cs:20,1.11435409478464)
--(axis cs:21,1.11166657036094)
--(axis cs:22,1.13487294132144)
--(axis cs:23,1.18028424896903)
--(axis cs:24,0.962605307887711)
--(axis cs:25,1.1519558547035)
--(axis cs:26,0.875547866467509)
--(axis cs:27,1.1875)
--(axis cs:28,1)
--(axis cs:29,1)
--(axis cs:29,1)
--(axis cs:29,1)
--(axis cs:28,1)
--(axis cs:27,0.4375)
--(axis cs:26,0.0827854668658241)
--(axis cs:25,0.598044145296501)
--(axis cs:24,0.122216120683718)
--(axis cs:23,0.48043003674526)
--(axis cs:22,0.343575334540633)
--(axis cs:21,0.263333429639057)
--(axis cs:20,0.323825253041447)
--(axis cs:19,0.377984354685616)
--(axis cs:18,0.558368450212118)
--(axis cs:17,0.590008567620741)
--(axis cs:16,0.741533035250637)
--(axis cs:15,0.800338630216506)
--(axis cs:14,0.864125307775084)
--(axis cs:13,0.891253106729874)
--(axis cs:12,0.918468925408575)
--(axis cs:11,0.93215742020194)
--(axis cs:10,0.94403866017459)
--(axis cs:9,0.967994842816118)
--(axis cs:8,0.970613180980378)
--(axis cs:7,0.980212272987793)
--(axis cs:6,0.988097601317088)
--(axis cs:5,0.995631819434418)
--(axis cs:4,0.976788516354719)
--(axis cs:3,0.965072373033894)
--(axis cs:2,1)
--(axis cs:1,-0.00200819665525429)
--(axis cs:0,0)
--cycle;

\path [draw=color1, fill=color1, opacity=0.2]
(axis cs:0,0)
--(axis cs:0,0)
--(axis cs:1,1.10325497937786)
--(axis cs:2,1.0202965228778)
--(axis cs:3,1.06990107887178)
--(axis cs:4,1.0542755367261)
--(axis cs:5,1.04156023599136)
--(axis cs:6,1.07544927445253)
--(axis cs:7,1.05677464999174)
--(axis cs:8,1.04082923850138)
--(axis cs:9,1.04552860138747)
--(axis cs:10,1.05171090319987)
--(axis cs:11,1.05614790137085)
--(axis cs:12,1.06959951215792)
--(axis cs:13,1.0767832157879)
--(axis cs:14,1.08999311626253)
--(axis cs:15,1.09684441580454)
--(axis cs:16,1.12284718161622)
--(axis cs:17,1.14323202229438)
--(axis cs:18,1.16682599744892)
--(axis cs:19,1.17330227403404)
--(axis cs:20,1.15874342902001)
--(axis cs:21,1.16591987448415)
--(axis cs:22,1.12499282592589)
--(axis cs:23,1.14536118799506)
--(axis cs:24,1.12278594114071)
--(axis cs:25,1.08190972086578)
--(axis cs:26,1.1275461957396)
--(axis cs:27,0.818847145934861)
--(axis cs:28,1)
--(axis cs:29,0.015625)
--(axis cs:30,1)
--(axis cs:30,1)
--(axis cs:30,1)
--(axis cs:29,0.015625)
--(axis cs:28,1)
--(axis cs:27,-0.0636388126015277)
--(axis cs:26,0.594676026482627)
--(axis cs:25,0.19309027913422)
--(axis cs:24,0.303255725525959)
--(axis cs:23,0.311670062004936)
--(axis cs:22,0.284729396296329)
--(axis cs:21,0.480526203947226)
--(axis cs:20,0.392727159215288)
--(axis cs:19,0.602595516021205)
--(axis cs:18,0.602433594877223)
--(axis cs:17,0.75148770334422)
--(axis cs:16,0.811292603330021)
--(axis cs:15,0.861700376781513)
--(axis cs:14,0.874521733795143)
--(axis cs:13,0.897410200075374)
--(axis cs:12,0.908648151393483)
--(axis cs:11,0.930475596967735)
--(axis cs:10,0.936583143943557)
--(axis cs:9,0.945568237317243)
--(axis cs:8,0.951659788288392)
--(axis cs:7,0.926234146339366)
--(axis cs:6,0.889104363957077)
--(axis cs:5,0.950262719747575)
--(axis cs:4,0.930045543543963)
--(axis cs:3,0.900916625554329)
--(axis cs:2,0.977903027009672)
--(axis cs:1,0.733654247928971)
--(axis cs:0,0)
--cycle;

\addplot [thick, color0]
table {%
0 0
1 0.0739462209302326
2 1
3 0.998068267066767
4 0.999062265566392
5 0.999962487808538
6 0.999756170755495
7 0.99932478055368
8 0.998611319621678
9 0.998317542672381
10 0.995474578194749
11 0.993636926711224
12 0.990975345326052
13 0.985291033647125
14 0.979527717472676
15 0.962257764948084
16 0.94338339640492
17 0.8794140625
18 0.862142857142857
19 0.759506777108434
20 0.719089673913043
21 0.6875
22 0.739224137931034
23 0.830357142857143
24 0.542410714285714
25 0.875
26 0.479166666666667
27 0.8125
28 1
29 1
};
\addlegendentry{ovap-strict}
\addplot [thick, color0, opacity=0.2, dashed, forget plot]
table {%
0 0
1 0.149900638515719
2 1
3 1.03106416109964
4 1.02133601477806
5 1.00429315618266
6 1.0114147401939
7 1.01843728811957
8 1.02660945826298
9 1.02864024252864
10 1.04691049621491
11 1.05511643322051
12 1.06348176524353
13 1.07932896056438
14 1.09493012717027
15 1.12417689967966
16 1.1452337575592
17 1.16881955737926
18 1.1659172640736
19 1.14102919953125
20 1.11435409478464
21 1.11166657036094
22 1.13487294132144
23 1.18028424896903
24 0.962605307887711
25 1.1519558547035
26 0.875547866467509
27 1.1875
28 1
29 1
};
\addplot [thick, color0, opacity=0.2, dashed, forget plot]
table {%
0 0
1 -0.00200819665525429
2 1
3 0.965072373033894
4 0.976788516354719
5 0.995631819434418
6 0.988097601317088
7 0.980212272987793
8 0.970613180980378
9 0.967994842816118
10 0.94403866017459
11 0.93215742020194
12 0.918468925408575
13 0.891253106729874
14 0.864125307775084
15 0.800338630216506
16 0.741533035250637
17 0.590008567620741
18 0.558368450212118
19 0.377984354685616
20 0.323825253041447
21 0.263333429639057
22 0.343575334540633
23 0.48043003674526
24 0.122216120683718
25 0.598044145296501
26 0.0827854668658241
27 0.4375
28 1
29 1
};
\addplot [thick, color1]
table {%
0 0
1 0.918454613653413
2 0.999099774943736
3 0.985408852213053
4 0.992160540135034
5 0.995911477869467
6 0.982276819204801
7 0.991504398165551
8 0.996244513394884
9 0.995548419352357
10 0.994147023571714
11 0.993311749169294
12 0.989123831775701
13 0.987096707931639
14 0.982257425028835
15 0.979272396293027
16 0.967069892473118
17 0.9473598628193
18 0.88462979616307
19 0.887948895027624
20 0.775735294117647
21 0.823223039215686
22 0.704861111111111
23 0.728515625
24 0.713020833333333
25 0.6375
26 0.861111111111111
27 0.377604166666667
28 1
29 0.015625
30 1
};
\addlegendentry{ovap-loose}
\addplot [thick, color1, opacity=0.2, dashed, forget plot]
table {%
0 0
1 1.10325497937786
2 1.0202965228778
3 1.06990107887178
4 1.0542755367261
5 1.04156023599136
6 1.07544927445253
7 1.05677464999174
8 1.04082923850138
9 1.04552860138747
10 1.05171090319987
11 1.05614790137085
12 1.06959951215792
13 1.0767832157879
14 1.08999311626253
15 1.09684441580454
16 1.12284718161622
17 1.14323202229438
18 1.16682599744892
19 1.17330227403404
20 1.15874342902001
21 1.16591987448415
22 1.12499282592589
23 1.14536118799506
24 1.12278594114071
25 1.08190972086578
26 1.1275461957396
27 0.818847145934861
28 1
29 0.015625
30 1
};
\addplot [thick, color1, opacity=0.2, dashed, forget plot]
table {%
0 0
1 0.733654247928971
2 0.977903027009672
3 0.900916625554329
4 0.930045543543963
5 0.950262719747575
6 0.889104363957077
7 0.926234146339366
8 0.951659788288392
9 0.945568237317243
10 0.936583143943557
11 0.930475596967735
12 0.908648151393483
13 0.897410200075374
14 0.874521733795143
15 0.861700376781513
16 0.811292603330021
17 0.75148770334422
18 0.602433594877223
19 0.602595516021205
20 0.392727159215288
21 0.480526203947226
22 0.284729396296329
23 0.311670062004936
24 0.303255725525959
25 0.19309027913422
26 0.594676026482627
27 -0.0636388126015277
28 1
29 0.015625
30 1
};
\end{axis}
\end{tikzpicture}
    \caption{Letting the OvA-Primal initial vector be the result of training
    the all-zero-label problem until convergence (strict) results in much
    worse performance than doing early stopping (loose). We suspect that this
    is because strict training brings the weight vectors into a regime that is
    qualitatively similar to the $\text{bias}=-1$ setting: All training
    instances will predict a negative label with a margin very close to $1$.
    Even though this results in much lower initial loss (left), it also leads
    to a useless first iteration in which little progress is made (right).}
    \label{fig:ovap:obj-and-ls}
\end{figure}

\begin{figure}[H]
    \input{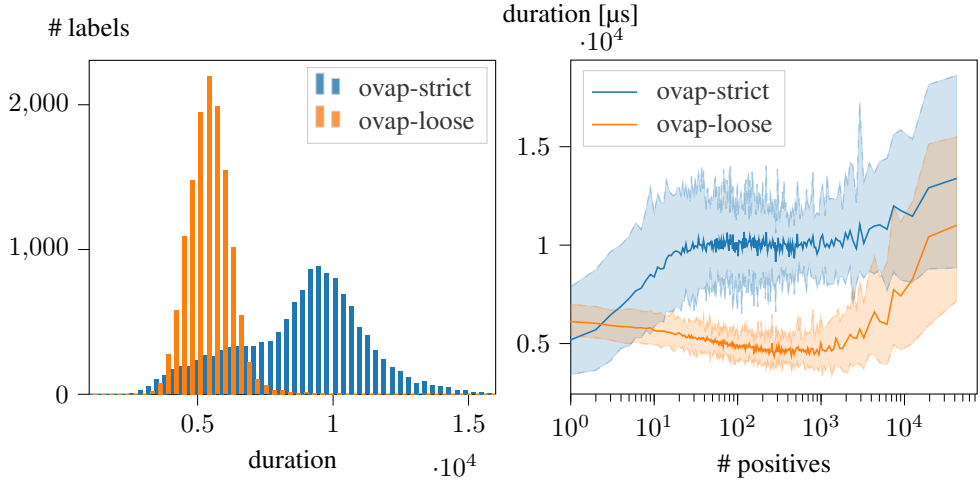}
    \caption{Histogram (left) and average with 95\% confidence interval in
    dependence of the number of positive instance (right) of the training
    time, measured in milliseconds. Overall, the \texttt{loose} training
    configuration is much faster, though for extreme tail labels with less
    than 3 positive instances, \texttt{strict} actually performs slightly
    better. }
    \label{fig:ovap:total-duration}
\end{figure}

The two figures in this section justify our choice of using a loose stopping
criterion for the \texttt{ovap} baseline in the main part of the paper. A more
strict criterion leads to lower initial loss, but results in weights
unfavourable for the Newton optimizer \autoref{fig:ovap:obj-and-ls}. As a
consequence, the \texttt{strict} convergence initial vector leads to overall
slower training, as shown in \autoref{fig:ovap:total-duration}.

\subsection{Choice of hyperparameters for \texttt{aop}}
\begin{figure}[htb]
    \input{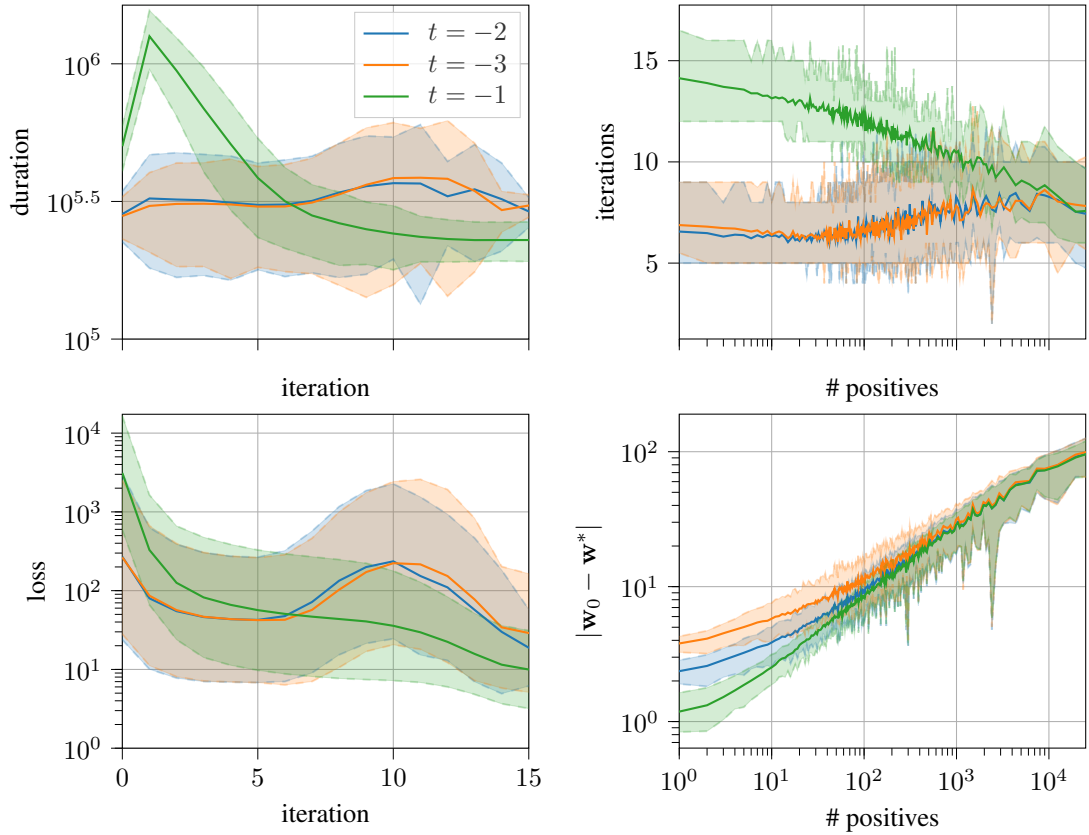}
    \caption{Average duration of the iterations (top left, in µs), and number
    of iterations for different numbers of positive instances (top right). The
    bottom left graph shows the development of the objective function. At
    first glance, this seems paradoxical, as the algorithm is designed so that
    for each binary problem, the loss strictly decreases over the iterations.
    The apparent increase is caused by the fact that the sample of binary
    problems shrinks for later iterations, as the labels for which the loss is
    already low before will terminate the optimization. Finally, the bottom
    right graph shows the Euclidian distance between the initial weight vector
    and the approximately optimal weight vector for which the minimization is
    terminated. The shaded area marks the 95\% quantile.}
    \label{fig:s-and-t:iters}
\end{figure}

As argued in the main text, in order to achieve sparsity in the Hessian
calculations, having the average of the negative instances be classified very
strongly as a negative would be beneficial. Interestingly,
\autoref{fig:s-and-t:iters} shows not only the expected speedup in per-iteration
time, but also that $t=-1$ requires significantly more iterations than the
other two settings.

This is not too surprising, given that we expect that the large number of
negative instances fill a larger proportion of the space than the few
positives, so the separating hyperplane should be closer to the mean of the
positives. One reason for this could be that $t=-1$ induces a higher initial
loss than the other two settings, as shown in the bottom left graph. However,
if we measure the distance that the optimization procedure covers,
\ie the distance between the initial and final weight vector, it turns out
that $t=-1$ actually starts closest. 

\subsection{Additional graphs for comparing the methods}
\begin{figure}[htb]
    \input{supplementarty/aop/Iterations}
    \caption{Number of Newton optimization steps required for convergence. On
    the left, the data is shown as a histogram, with the bins corresponding to
    the number of binary problems that required the corresponding number of
    steps. On the right, the average number of steps, and the 95\% quantiles,
    are plotted as a function of the number of positive instances.
    \label{fig:aop:iterations}}
\end{figure}
\begin{figure}[htb]
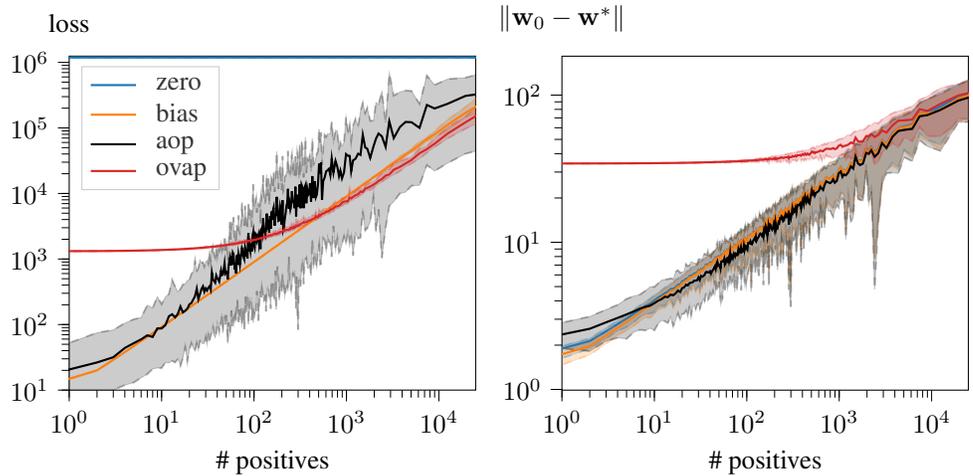

    \input{supplementarty/aop/InitLoss}
    \input{supplementarty/aop/TrainingShift}
    \caption{Loss of the initial weight vector (left) and distance between
    initial and final vector (right). The shaded area shows 95\% quantiles.
    For all except \texttt{zero} initialization, loss and distance are much lower
    for tail labels than for head labels.}
    \label{fig:aop:loss-shift}
\end{figure}

As shown in \autoref{fig:aop:iterations}, using the \texttt{aop}
initialization drastically reduces the number of iterations needed for
convergence for tail labels. Whereas for \texttt{zero} and \texttt{ovap}
initialization the number of iterations increases as the number of positive
instances decreases whereas for \texttt{bias} and \texttt{aop} initialization
it remains almost constant. This cannot be explained by looking only at the
distance the optimization algorithm has to travel, neither in terms of the
initial loss value, nor in terms of the distance between initial and final
weight vector. In \autoref{fig:aop:loss-shift} we see that these values are
much lower for tail labels (except for \texttt{zero} init). 

\begin{figure}[htb]
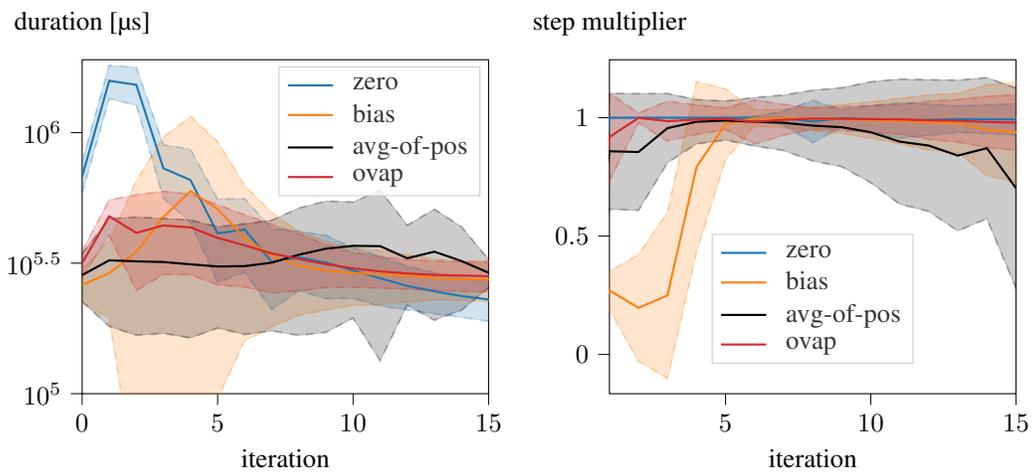

    \input{supplementarty/aop/IterDuration}
    \input{supplementarty/aop/LineSearch}
    \caption{Duration of a single iteration (left), and line search step
    multiplier (right). The \texttt{aop} initialization manages to have the
    early iterations as fast as the later ones, while still working much
    better in terms of the quadratic approximation than the \texttt{bias}
    initialization. The decrease of the step multiplier for late iterations is
    not relevant, because as \autoref{fig:aop:iterations} shows most of the
    binary problems will have finished by iteration 10.}
    \label{fig:aop:iter-dur}
\end{figure}

In \autoref{fig:aop:iter-dur} the duration of a single iteration, and the step
size multiplier of the backtracking line search are presented.

\clearpage
\subsection{Full Result Table}
\begin{table}[H]
\centering
\caption{Durations (as given in main paper) and precision@\{1,3,5\} for
training different datasets with the given initialization methods. Since the
actual loss function is based on binary classification, we have also
calculated the precision and recall as averaged over the individal binary
problems. The metrics remain mostly unchanged for tf-idf data, but there is
significant fluctuation when \texttt{slice} features are used. The
\texttt{aop*} setting refers to a training run where $t=-3$ was used, which
turned out to be beneficial in the logistic loss setting.}
    \begin{filecontents*}{results.txt}
Dataset         Features    Method      P@1     P@3     P@5     Prec    Rec     Duration
Eurlex-4k       tf-idf      zero        82.8    70.3    58.8    80.3    37.6       11      
Eurlex-4k       tf-idf      bias        82.9    70.4    58.9    80.2    37.6        8
Eurlex-4k       tf-idf      aop         82.9    70.4    58.9    80.3    37.6        6
Eurlex-4k       tf-idf      ovap        82.9    70.4    58.9    80.3    37.6        7
Amazoncat-13k   tf-idf      zero        93.8    79.1    64.1    84.4    58.5     1049
Amazoncat-13k   tf-idf      bias        93.8    79.1    64.1    84.5    58.4      557
Amazoncat-13k   tf-idf      aop         93.7    79.1    64.1    84.4    58.4      348
Amazoncat-13k   tf-idf      ovap        93.8    79.1    64.1    84.4    58.5      640
Amazoncat-14k   tf-idf      zero        89.9    69.6    54.9    87.1    66.8     4563
Amazoncat-14k   tf-idf      bias        89.8    69.6    54.9    87.1    66.7     1790
Amazoncat-14k   tf-idf      aop         89.8    69.6    54.9    87.2    66.6     1217
Amazoncat-14k   tf-idf      ovap        89.8    69.6    54.9    87.2    66.6     2497            
WikiLSHTC       tf-idf      zero        64.2    42.5    31.5    84.4    19.3    66660
WikiLSHTC       tf-idf      bias        64.1    42.5    31.5    84.5    19.3    28769
WikiLSHTC       tf-idf      aop         64.2    42.5    31.6    84.4    19.4    18628
WikiLSHTC       tf-idf      ovap        64.2    42.5    31.5    84.4    19.3    41278
WikiTitles-500k tf-idf      zero        39.3    20.8    14.7    66.5     5.3    17559
WikiTitles-500k tf-idf      bias        39.3    20.8    14.7    66.6     5.3     5598
WikiTitles-500k tf-idf      aop         39.3    20.8    14.7    66.7     5.3     4310
WikiTitles-500k tf-idf      ovap        39.3    20.8    14.7    66.5     5.3    18920
Amazon-670k     tf-idf      zero        44.8    39.7    36.2    79.3    18.7    19026
Amazon-670k     tf-idf      bias        44.8    39.7    36.1    79.3    18.7     8926
Amazon-670k     tf-idf      aop         44.8    39.7    36.2    79.3    18.7     5045
Amazon-670k     tf-idf      ovap        44.7    39.8    36.1    79.3    18.7    11840
Amazon-670k     slice       zero        35.6    32.1    29.4    74.6    23.0    46945
Amazon-670k     slice       bias        34.2    31.0    28.5    69.4    22.4    22562
Amazon-670k     slice       aop         35.6    32.1    29.5    73.5    23.4    19272
Amazon-670k     slice       ovap        35.6    32.0    29.4    74.5    23.0    29350
Amazoncat-13k   logistic    zero        93.2    77.5    61.9    85.8    47.2    1758
Amazoncat-13k   logistic    bias        93.2    77.5    61.8    85.8    47.1    1517
Amazoncat-13k   logistic    aop         93.2    77.5    61.8    85.8    47.1    1508
Amazoncat-13k   logistic    ovap        93.1    77.5    61.8    85.8    47.1    1214
Wiki10-31k      tf-idf      zero        84.1    74.7    66.0    75.7    13.2    239
Wiki10-31k      tf-idf      bias        84.2    74.8    66.0    75.7    13.2    215
Wiki10-31k      tf-idf      aop         84.2    74.7    66.0    75.6    13.2    141
Wiki10-31k      tf-idf      ovap        84.2    74.7    66.0    75.6    13.2    189
Eurlex-4k       slice       zero        77.2    63.4    51.3    72.8    36.3    8
Eurlex-4k       slice       bias        75.3    62.1    50.2    76.0    26.4    4
Eurlex-4k       slice       aop         76.6    63.0    51.4    67.7    39.1    3
Eurlex-4k       slice       ovap        76.9    63.5    51.4    72.5    36.4    4
Amazoncat-13k   logistic    aop*        93.0    77.4    61.7    85.7    47.1    1410
Amazon-3M       tf-idf      aop         47.3    44.4    42.3    65.1    14.8    69774
\end{filecontents*}
\pgfplotstabletypeset[ every head row/.style={before row={\toprule}, after
    row=\midrule,}, every last row/.style={after row=\bottomrule},
    columns/Dataset/.style={string type, column type=l},
    columns/Features/.style={string type, column type=l, column
    name={Setting}}, columns/Method/.style={string type, column type=l},
    columns/Duration/.style={divide by=60, fixed, fixed zerofill, precision=2,
    column type=r}, columns/P@1/.style={fixed zerofill, precision=1},
    columns/P@3/.style={fixed zerofill, precision=1},
    columns/P@5/.style={fixed zerofill, precision=1},
    columns/Prec/.style={fixed zerofill, precision=1},
    columns/Rec/.style={fixed zerofill, precision=1}, empty cells
    with={\textemdash}, every nth row={4}{before row=\midrule} ] {results.txt}
    \label{tabel:full-results}
\end{table}

\end{document}